\documentclass[9pt,technote]{IEEEtran}

\usepackage{graphics} 
\usepackage{epsfig} 
\usepackage{mathptmx} 
\usepackage{times} 
\usepackage{amsmath} 
\usepackage{amssymb}  
\usepackage{multirow}
\usepackage{url}
\usepackage{comment}
\usepackage{siunitx}
\usepackage{wrapfig}
\DeclareMathOperator*{\argmax}{arg\,max} 
\usepackage{balance}

\title{\LARGE \bf
	A Holistic Visual Place Recognition Approach using Lightweight CNNs for Significant ViewPoint and Appearance Changes 
}

\author{Ahmad Khaliq$^{1}$, Shoaib Ehsan$^{1}$, Zetao Chen$^{2}$, Michael Milford$^{3}$ and Klaus McDonald-Maier$^{1}$
	\thanks{This work is supported by the UK Engineering and Physical Sciences Research Council through grants EP/R02572X/1 and EP/P017487/1.}
	\thanks{$^{1}$Authors are with Embedded and Intelligent System Lab in School of Computer Science and Electronic Engineering,
   	 University of Essex, Colchester, United Kingdom
   	 {\tt\small \{ahmad.khaliq,sehsan,kdm\}@essex.ac.uk},{\tt\small\ ahmedest61@hotmail.com}.}%
	\thanks{$^{2}$Zetao Chen is with the Vision for Robotics Lab, ETH Zurich, Zurich 8092, Switzerland
		{\tt\small chenze@ethz.ch}.}%
	\thanks{$^{2}$Michael Milford is with the Australian Centre for Robotic Vision and School of Electrical Engineering and Computer Science, Queensland University of Technology, Brisbane, Australia
   	 {\tt\small michael.milford@qut.edu.au}.}%

}

\begin{document}

	\maketitle
	\thispagestyle{empty}
	\pagestyle{empty}

	\begin{abstract}
		
		
		
		This paper presents a lightweight visual place recognition approach, capable of achieving high performance with low computational cost, and feasible for mobile robotics under significant viewpoint and appearance changes. Results on several benchmark datasets confirm an average boost of 13\% in accuracy, and 12x average speedup relative to state-of-the-art methods.

	\end{abstract}
	
	
	\vspace{-2mm}
	\section{INTRODUCTION}
	
	Given a query image, an image retrieval system aims to retrieve all images from a large database that contain similar objects as in the query image. Visual Place Recognition (VPR) can also be interpreted as an image retrieval system that tries to recognize a place by matching it with the places from the stored database \cite{lowry2016visual}. A place database is a simplest way to represent a particular environment where appearance based information is stored as an image with no pose related data. However, other VPR techniques use topological maps which contain relative information about the places in an environment (can be an ordered collection of images) and metric maps which are even more accurate in terms of absolute scale of the environment (such as distance, landmark position) but difficult to build and maintain. Two image matching techniques; 1) single image and 2) sequence of images are employed by the VPR community. This paper focuses on database-centric place remembering approach coupled with single image matching, thus, place recognition is solely based on appearance similarity and image retrieval techniques are applicable \cite{zheng2018sift}. 
	
	\begin{figure}[t]
		\centering
		\vspace{-2mm}
	    \includegraphics[width=0.88\linewidth, height=1\linewidth,keepaspectratio]{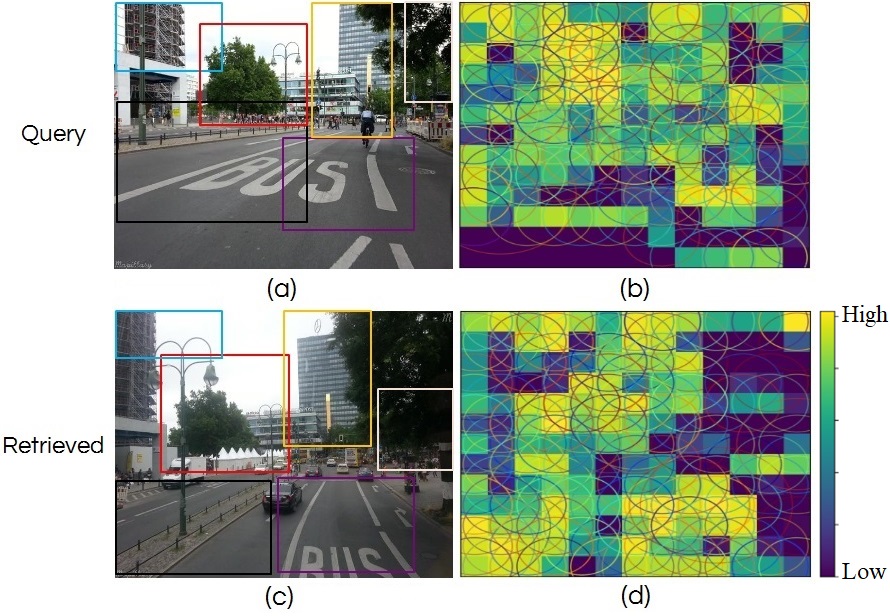}
		\caption{For a query image (a), the proposed Region-VLAD approach successfully retrieves the correct image (c) from a stored image database under significant viewpoint- and condition-variation. (b) and (d) represent their CNN based meaningful regions identified by our proposed methodology.}
		\label{figure:topMatchedImg}
		\vspace{-5mm}
	\end{figure}
	
	As with a range of other computer vision applications, Convolutional Neural Networks (CNNs) have shown promising results for VPR and managed to shift the focus from traditional hand-crafted feature descriptors \cite{bay2006surf}\cite{lowe2004distinctive} to CNNs \cite{chen2014convolutional}\cite{sunderhauf2015place}\cite{chen2017only}. Using a pre-trained CNN for VPR, there are three standard approaches to produce a compact image representation: (a) the entire image is directly fed into the CNN and responses from convolutional layers are extracted \cite{chen2014convolutional}; (b) CNN is applied on user-defined regions of the image and prominent activations are pooled from the layers representing those regions \cite{sunderhauf2015place}; (c) the entire image is fed into the CNN and salient regions are identified by directly extracting distinguishing patterns based on convolutional layers responses \cite{chen2017only}\cite{chen2018learning}. Generally, global image representations retrieved from category (a) are not robust against strong viewpoint variations and partial occlusion. Image representations emerging from category (b) usually handle viewpoint changes better but are computation intensive. On the other hand, image representations resulting from category (c) address both the appearance and viewpoint variations. In this paper, we focus on category (c).       

	The work by \cite{chen2017only} and \cite{chen2018learning} are considered as state-of-the-arts in identifying prominent regions by directly extracting unique patterns based on convolutional layers' responses. 
	Chen \textit{et al.} in \cite{chen2017only} used VGG-16 network \cite{simonyan2014very} pre-trained on ImageNet \cite{krizhevsky2012imagenet} and used late convolutional activations for regions identification. For regions-based feature encoding, $10k$ bag-of-words (BoW) \cite{sivic2003video} codebook is employed. The system is tested on five benchmark place recognition datasets with AUC-PR curves \cite{hanley1982meaning} as the evaluation metric. It claims to outperform FABMAP \cite{cummins2008fab}, SEQSLAM \cite{milford2012seqslam} and other image retrieval pooling techniques including Cross-Pooling \cite{liu2017cross}, Sum/Average-Pooling \cite{babenko2015aggregating} and Max-Pooling \cite{tolias2015particular}.
	
	Despite its good AUC-PR performance, the method proposed in \cite{chen2017only} has some shortcomings. A common strategy for improving CNN accuracy is to make it deep by adding more layers (provided sufficient data and strong 
	regularization). However, increasing network size means increased computation and using more memory both at time of training and testing (such as, for storing outputs of intermediate layers and for storing parameters) which is not ideal for resource-constrained robots that are usually battery-operated. Using $10k$ BoW dictionary for regions-based feature encoding (extracted from late convolutional layers of deep VGG-16) followed up with their cross-matching thus degrades the real-time performance. Secondly, employment of object-centric deep VGG-16 model results in system attempting to put more emphasis on objects rather than the place itself. This reflects on the regions-based pooled feature and leads to failure cases. Also, the regional approach proposed in \cite{chen2017only} hinders the identification of individual static place-centric regions that can be more effective under condition and viewpoint variations. 
	
	To bridge these research gaps, this paper proposes a holistic approach targeted for a CNN architecture comprising a small number of layers pre-trained on a scene-centric \cite{zhou2017places} image databases to reduce the memory and computational costs. The proposed method detects novel CNN-based regional features and combines them with VLAD \cite{jegou2010aggregating} adapted specifically for the VPR problem.
	The motivation behind employing VLAD comes from its better performance in various CNN-based image retrieval tasks utilizing a smaller visual word dictionary \cite{jegou2010aggregating}\cite{farandjelovic2013all} compared to BoW \cite{sivic2003video}. To the best of our knowledge, this is the first work that
	combines novel lightweight CNN-based regional features with VLAD encoding adapted
	for computation-efficient VPR under changing environment. 
	
	As opposed to \cite{chen2017only} which uses object-centric VGG-16 architecture and employs a cross-convolution based regional extraction approach (resembles \cite{liu2017cross}), the proposed VPR technique is particularly different both in identification and extraction of regional features (discussed in detail, in section \ref{section:ROI}). 
	The presented approach in this paper showcases enhanced accuracy by employing middle convolutional layer of the CNN architecture comprising of small number of layers. Evaluation on several viewpoint- and condition-variant benchmark place recognition datasets shows an average performance boost of 13\% over state-of-the-art VPR algorithms in-terms of AUC computed under Precision-Recall curves. In Figure \ref{figure:topMatchedImg}, for a query image (a), our proposed system retrieved image (c) from the stored database. (b) and (d) highlight the salient regions which our proposed methodology identified under strong visual changes.  
	
	The rest of the paper is organized as follows. Section II provides the related work for VPR and other image retrieval tasks. 
	In Section III, the proposed methodology is presented in detail. Section IV illustrates the implementation details and performance evaluation of the proposed VPR framework on several benchmark datasets. Section V presents the conclusion.

	\vspace{-3.5mm}
	\section{Literature Review}
	
	This section provides an overview of major developments in VPR under simultaneous viewpoint and appearance changes using hand-crafted features and CNN-based features. Other image retrieval tasks with their feature extracting and encoding approaches are further discussed and differentiated from VPR based image retrieval tasks.  
	
	FAB-MAP \cite{cummins2008fab} is the first work that used handcrafted SURF feature descriptors combined with BoW encoding for VPR. It demonstrated robustness under viewpoint changes by taking advantage of the in-variance properties of SURF. Another sequence-based image matching technique, SEQSLAM \cite{milford2012seqslam} has shown remarkable performance under severe appearance changes. However, it is unable to deal with simultaneous condition- and viewpoint-variation.
	
	The first CNN-based VPR system is introduced in \cite{chen2014convolutional}, which is followed by \cite{sunderhauf2015performance}, \cite{sunderhauf2015place} and \cite{panphattarasap2016visual}. Chen \textit{et al.} in  \cite{chen2014convolutional} used Overfeat \cite{sermanet2013overfeat} trained on ImageNet. Eynsham \cite{cummins2008fab} and QUT datasets with multiple traverses of the same route under environmental changes are used for benchmarking. Using the Euclidean distance on the pooled layers' responses, test images are matched against the reference images. On the other hand, authors in \cite{panphattarasap2016visual} and \cite{sunderhauf2015place} used landmark-based approaches coupled with the pre-trained CNN models. Chen \textit{et al.} in \cite{chen2017deep} introduced two CNN models for the specific task of VPR (named AMOSNet and HybridNet) which trained and fine-tuned the object-centric CaffeNet \cite{krizhevsky2012imagenet} on a $2.5$ million Specific PlacEs Dataset (SPED). The place-recognition centric SPED  consists of thousands of places with severe-condition variance among the same places over different times of the year. The results showed that with Spatial Pyramidal Pooling (SPP) employed on middle and late convolutional layers, HybridNet outperformed AMOSNet, CaffeNet and PlaceNet on four publicly available datasets exhibiting strong appearance and moderate viewpoint changes \cite{chen2017deep}. 
	
	Chen \textit{et al.} in \cite{chen2017only} presented a VPR approach that identifies pivotal landmarks by directly extracting prominent patterns based on responses of late convolutional layers of deep object-centric VGG-16 model. 
	Recently, Chen \textit{et al.} in \cite{chen2018learning} introduced a context-flexible attention model and combined it with a pre-trained object-centric VGG-16 fine-tuned on SPED \cite{chen2017deep} to learn more powerful condition-invariant regional features. The system has shown state-of-the-art performance on severe condition-variant datasets. 
	However, the efficiency of the framework may be compromised if there is a simultaneous strong viewpoint and condition variations. Moreover, performance and efficient resource usage become two important aspects to be looked upon in real-life robotic VPR applications. 
	
	Image retrieval tasks which either rely on handcrafted features, such as, local SIFT and SURF features \cite{bay2006surf}\cite{lowe2004distinctive} or combining these with convolutional and fully connected layers of deep/shallow CNNs \cite{zheng2018sift}\cite{yue2015exploiting}\cite{chen2014convolutional}, Bag-of-Words (BoW) or Support Vector Machine (SVM) \cite{agarap2018neural} are employed for classification, detection and recognition \cite{tolias2015particular}\cite{liu2017cross} purposes. 
	As an alternative for BoW feature encoding scheme, several other approaches including Fisher vector \cite{sanchez2013image} and Vector of Locally aggregated descriptor (VLAD) have shown promising results with smaller visual words vocabularies \cite{jegou2010aggregating}. To perform instance level image retrieval where objects from the same category are to be separated, 
	Yue-Hei Ng \textit{et al.} in \cite{yue2015exploiting} suggested to combine rich spatial middle convolutional layers' features with VLAD encoding. 
	Kim \textit{et al.} in \cite{jin2015predicting} have used MSER \cite{matas2004robust} for regions identification, coupled with SIFT feature description within the identified regions and described each region/bundle as a fix sized VLAD, named as PBVLAD. 
	2D-based localization methods generally offer efficient database management at low accuracy cost whereas 3D-based techniques are computationally complex but more reliable in localization. Sattler \textit{et al.} in \cite{sattler2017large} refute this notation by combining 2D-based approaches with SfM-based post-processing and shown better performances then structure-based methods. However, such post processing takes significant longer run-times which is out of scope of this work since our proposed VPR system works like a 2D-based framework with an aim to improve the retrieval performance while reducing the computation complexities.  
	
	With the advent of several feature pooling techniques including Sum-Pooling \cite{babenko2015aggregating}, Max-Pooling \cite{tolias2015particular}, Spatial Max-Pooling \cite{jaderberg2015spatial} and Cross-Pooling  \cite{liu2017cross} employed in deep CNNs have demonstrated performance boost in tasks requiring image classification/recognition and object detection/retrieval \cite{tolias2015particular}\cite{liu2017cross}. All these pooling approaches process the convolutional layers' feature maps as a whole to pick prominent patterns, and with images focus on fewer objects make feature maps sparse in nature and finding single region of interest becomes relatively easier. However, such image retrieval tasks are different in nature from the VPR systems where recognizing a place which undergoes diverse changes due to illumination, winter-summer transition or viewpoint variance added by different capturing angles is quite challenging because appearance of the place changes and makes it difficult to identify the common regions. For VPR, when such external tasks based pre-trained CNNs \cite{krizhevsky2012imagenet} are integrated with the above mentioned feature pooling techniques, the convolutional layers feature maps focus on the trained objects such as vehicles, pedestrians and other time varying objects which are not suitable for place recognition \cite{chen2017only}. Therefore, it is still questionable for a generic VPR system to efficiently deal with simultaneous viewpoint and condition variations when employing CNN-based local features pre-trained on other image retrieval tasks. 

	Recently, Teichmann \textit{et al.} in \cite{teichmann2018detect} trained the landmark detectors \cite{razavian2016visual}\cite{tolias2015particular} with a newly introduced $1.2M$ Google Landmark dataset (GLD) containing $15k$ landmark categories (such as, buildings, monuments and bridges) annotated by human. 
	Noticing that not all the visual words get associated with the feature descriptors which results into many zero regional residuals, their proposed R-VLAD technique overcomes it by normalizing the regional residuals \cite{teichmann2018detect}.
	Precisely, it down-weights all the regional residuals 
	and stores a single aggregated regional descriptor per image. Custom landmark detectors including ASMK \cite{tolias2016image}, RMACB \cite{razavian2016visual}, RMAC \cite{tolias2015particular} and selective search \cite{tao2014locality} are incorporated for the regional search and coupled with R-VLAD on deep CNNs. 
	We can expect further boost in our proposed VPR framework with the integration of R-VLAD \cite{teichmann2018detect}.  
	Chen \textit{et al.} in \cite{chen2018learning} have shown that the state-of-the-art regions-based image retrieval techniques including Attentive Attention \cite{yu2017multi} and Fixed Context \cite{jin2017learned} are not generally efficient for VPR under strong visual changes.  

	\vspace{-2mm}
	\section{Proposed Technique}
\begin{figure}[h]
		\centering
		\vspace{-3mm}
		\includegraphics[width=1\linewidth, height=1\linewidth,keepaspectratio]{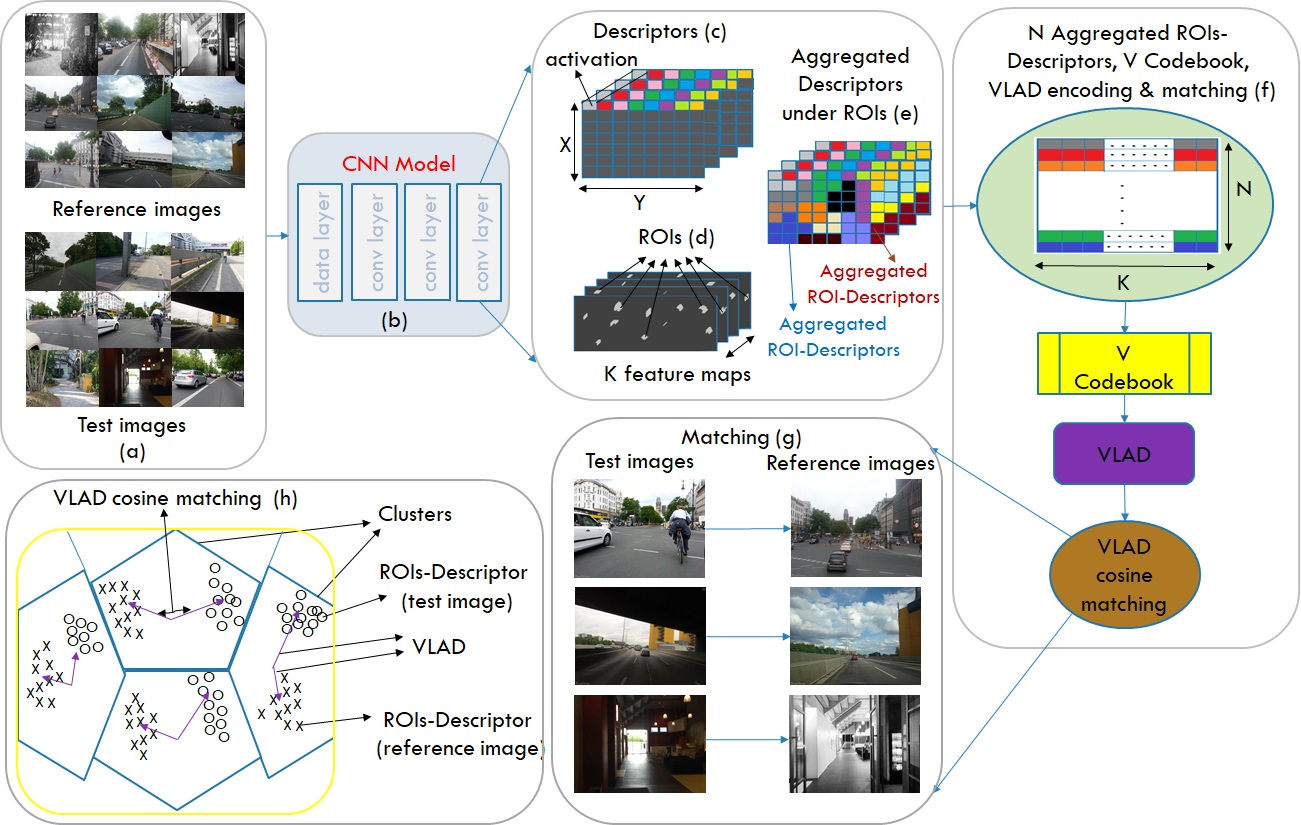}
		\caption{Workflow of the proposed VPR framework is shown here. Test/reference images are fed into the CNN model, Region-of-Interests (ROIs) are identified across all the feature maps of the convolution layer and their compact VLAD representation is stored for image matching. 
		}
		\label{figure:proposedMethod}
		\vspace{-1mm}
	\end{figure}

	In this section, the key steps of the proposed methodology are described in detail. It starts from the idea of stacking activations of feature maps for retrieving local descriptors, followed up with the identification of distinguishing regional patterns. It then illustrates the aggregation of local feature descriptors lying under those identified salient regions. Finally, it shows how to retrieve the compact VLAD representation using the extracted CNN-based regional features, later used for determining a match between two images. The workflow of the proposed methodology is shown in Figure \ref{figure:proposedMethod}. 
	
	\vspace{-6mm}
	\subsection{Stacking of Convolutional Layer Activations for making Descriptors}
	\label{featuresStacking}
	Given an image $I$ as an input to the CNN model, at a certain convolutional layer, the output is a $ 3D $ tensor $ M $ of $ X \times Y \times K $ dimensions. $ K $ denotes the number of feature maps, $X$ and $Y$ represent the width and height of feature map / channel. We can also interpret it as $M^k$ being a set of $ X \times Y $ activations / responses for $k^{th}$ feature map where $k = \{ 1,2,....,K\}$. 
	For $K$ feature maps in the convolution layer, we stack each activation at some certain spatial location into $ K $ dimensional local feature as shown with different colours in Figure \ref{figure:proposedMethod} (c). $D^L$ in (\ref{eq:1}) represents the  $K$ dimensional $d_l$ feature descriptors at $L^{th}$ convolutional layer of $m_c$ model.
	\begin{equation}D^L =\{d_l  \in  M^K\ \ \forall  \ l \in \{(i,j) \ | \ i=1, ...,X; j=1, ..., Y\}\} , \ L \in m_c\vspace{-3mm}\label{eq:1}\end{equation}
	
	\subsection{Identification of Regions of Interest}
	\label{section:ROI}
\begin{figure}[h]
		\centering
		\vspace{-3mm}
		\includegraphics[width=1\linewidth, height=1\linewidth,keepaspectratio]{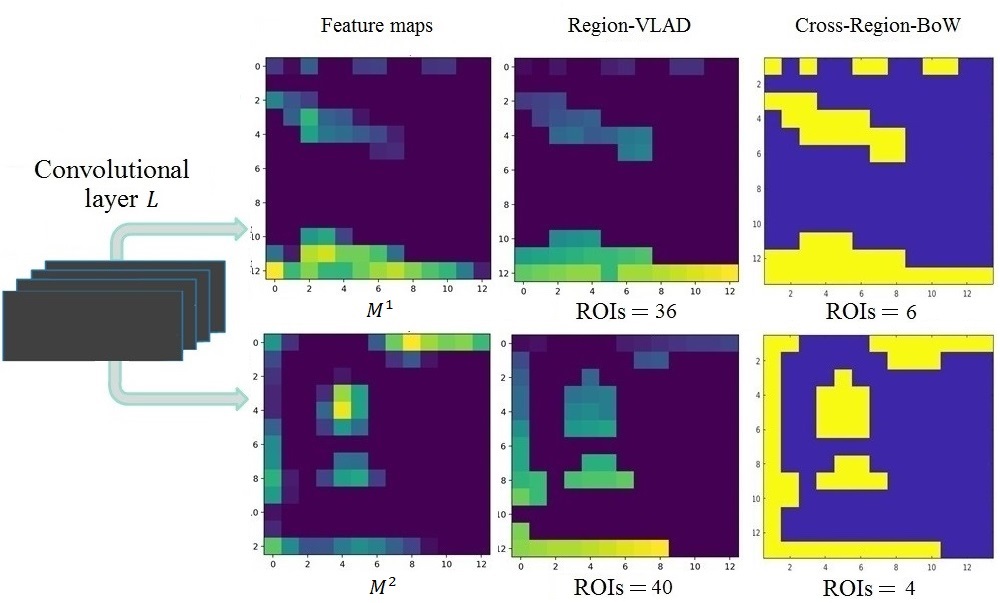}
		\caption{Employing two features maps $M^1$ and $M^2$, sample images of ROIs identified by Region-VLAD and Cross-Region-BoW \cite{chen2017only} are shown here. Note that feature maps ($1^{st}$ column) illustrate the intensities of $a$ activations. However, regardless of the intensity, each identified $G_h$ region per feature map for Region-VLAD ($2^{nd}$ column) is indicated with a different colour i.e. $36$ and $40$ coloured regions for feature map $M^1$ and $M^2$. For Cross-Region-BoW ($3^{rd}$ column), all the regions are denoted as yellow patterns i.e. $6$ and $4$ ROIs for $M^1$ and $M^2$ feature maps.   
		}
		\label{figure:labledROIs}
		\vspace{-2mm}	
	\end{figure}

	To extract region-based CNN features, the most prominent regions need to be identified. 
	Two or more activations are considered to be connected and represented as a region if they are neighbours and have approximately the same value. For $K$ feature maps, 
	each region is denoted by $G_h$, $\forall \ h \in \{1,...,H\} $ where $H$ is the total number of identified regions at $L^{th}$ convolution layer, visualized in Figure \ref{figure:proposedMethod}(d) or Figure \ref{figure:labledROIs}. 

	The mean energy of each $G_h$ region is calculated by averaging over all $a_h$ activations lying under the region. In (\ref{eq:2}), $a^f_h$ represents the $f^{th}$ activation lying under the $G_h$ region and $E^L$ denotes the calculated mean energies of $H$ regions. Based on the sorted $E^L$ energies, top $N$ energetic ROIs (with their bounding boxes) are picked in (\ref{eq:3}), denoted as $R^L$ novel regions at $L^{th}$ convolution layer. 
	
	\begin{equation}\vspace{-2mm} E^L = \{ \frac{1}{|G_h|} \ \sum\limits_f a^f_h,\ \forall \ a^f_h \in G_h\} \label{eq:2}\end{equation}

	\begin{equation}R^L = \{G_t \ \forall  \ t \in \{1,...,N\}\}\label{eq:3}\end{equation}
	
\begin{figure}[h]
		\centering
		\vspace{-3.25mm}
		\includegraphics[width=0.93\linewidth, height=1\linewidth,keepaspectratio]{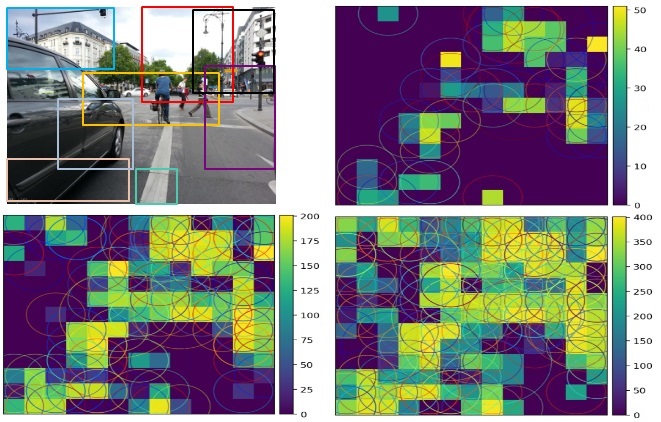}
		\caption{Sample images of top $N=\{50, 200, 400\}$ ROIs identified by our proposed approach at $L^{th}$ convolutional layer; CNN based identified regions put emphasis on static objects such as, buildings, trees and road signals.}
		\label{figure:50_rois}
		\vspace{-2.5mm}	
	\end{figure}

	Figure \ref{figure:50_rois} illustrates the top $N = \{50,200,400\}$ novel $R^L$ regions identified by our proposed regions-based VPR system. 
	Our novel CNN based identified regions strongly concentrate on the static objects including buildings, trees and road signals. $D^L$ local descriptors in (\ref{eq:1}) which fall under the bounding boxes of $R^L$ regions in (\ref{eq:3}), aggregated in (\ref{eq:4}) to retrieve CNN-based regional features. Intuitively, each regional feature is $1 \times K$ dimensional $f_t$ vector where $q$ be the $R^L_t$ region under which $D^L_q$ descriptors fall. For $N$ novel regions, (\ref{eq:5}) represents $N \times K$ dimensional $F^L$ region-based CNN features representing an image at $L^{th}$ convolutional layer (visually shown in Figure \ref{figure:proposedMethod} (e) / Figure \ref{figure:proposedMethod} (f)).  
	
	\vspace{-4mm}
	\begin{equation}f_t = \sum\limits_{q \in R^L_t} D^L_q \label{eq:4}\end{equation}

	\vspace{-3mm}
	\begin{equation}F^L = \{ f_t \ \forall  \ t \in \{1,...,N\}\}\vspace{-1mm}\label{eq:5}\end{equation}

	In comparison, authors in \cite{chen2017only} first identified regions, 
	calculated their mean energies and selected $N=200$ energetic regions. Precisely, $N$ regional activations at $L^{th}$ convolution layer were mapped onto the ${L-1}^{th}$ convolutional feature maps and aggregation of modified cross-mapped regions-based local descriptors at ${L-1}^{th}$ convolution layer was carried out for feature extraction.	
	Note, that depending upon the quantity of activations per ROI(s) at $L^{th}$ convolution layer and receptive field of the filter (e.g. $3\times3$, $5\times5$) for cross-mapping of $L^{th}$ convolution layer regions at ${L-1}^{th}$ layer, the bounding box (area) per cross-mapped regional feature varies for \cite{chen2017only}. 
	
\begin{figure}[t]
		\centering
		\vspace{-3mm}
		\includegraphics[width=1\linewidth, height=1\linewidth,keepaspectratio]{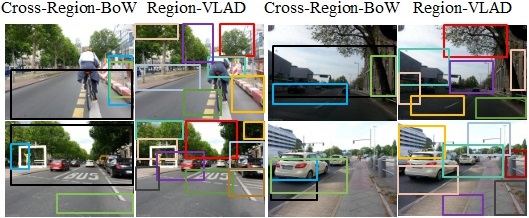}
		\caption{Sample images of ROIs identified with Cross-Region-BoW \cite{chen2017only} and Region-VLAD are shown here. Our regional approach subdivides each image into large number of most contributing regional blocks. 
		}
		\label{figure:R-BoW_Vs_R-VLAD}
		\vspace{-4.5mm}	
	\end{figure}

	Furthermore, Figure \ref{figure:labledROIs} illustrates that the identified ROIs from two feature maps ($M^1$ and $M^2$) at $L^{th}$ convolutional layer with Region-VLAD and Cross-Region-BoW \cite{chen2017only} are different in terms of quantity and size / activations per region(s). Thus, the computed regional mean energies of \cite{chen2017only} are different from the mean energies of regions identified by our approach. Our approach identifies $36$ and $40$ ROIs from feature map $M^1$ and $M^2$, shown with different colours. Later, based on their computed mean energies, top $N$ energetic regions are selected from $H$ identified regions at $L^{th}$ convolutional layer, as visualized in Figure \ref{figure:50_rois}. The 8-connected component-based regional approach in Cross-Region-BoW \cite{chen2017only} identifies $6$ and $4$ yellow coloured ROIs for feature map $M^1$ and $M^2$. As explained above, $N$ energetic regional feature extraction for \cite{chen2017only} is carried out by first selecting $N$ energetic regions at $L^{th}$ layer (Figure \ref{figure:labledROIs}) followed up with their mapping at ${L-1}^{th}$ convolution layer and aggregation of cross-mapped regions-based local descriptors at ${L-1}^{th}$ convolution layer (not shown in the figure). 
	Exemplars exhibiting the novel identified regions by Cross-Region-BoW \cite{chen2017only} and with our proposed Region-VLAD framework are shown in Figure \ref{figure:R-BoW_Vs_R-VLAD}. We observe that regional patterns covering more areas similar to \cite{chen2017only} hinder the identification of individual place-centric instances vital in recognizing places under changing conditions and viewpoints.
	
	\subsection{Regional Vocabulary and Extraction of VLAD for Image Matching}
	\label{section:vocabVLAD}
	Vector of Locally Aggregated Descriptor (VLAD) adopts K-means \cite{sivic2003video} based vector quantization, accumulates the residues of features quantized to each dictionary cluster and concatenates those accumulated vectors into a single feature representation. A separate dataset of $2.6k$ images is collected and afore-described regions-based feature extraction is employed for generating a regional vocabulary. To learn a diverse vocabulary, we employed $1125$ place-recognition centric images of $365$ places from Query247 \cite{torii201524} (taken at day, evening and night times). Other images include a benchmark place recognition dataset St.lucia \cite{chen2017deep} with $1k$ frames of two traverses captured in suburban environment at multiple times of the day. The left over images consist of multiple viewpoint- and condition-variant traverses of urban and suburban routes collected from \textit{Mapillary\footnote{https://www.mapillary.com/}} 
	(previously employed by \cite{sunderhauf2015place} and \cite{chen2017only} for capturing 
	place recognition datasets). K-means is employed for clustering the $2600 \times N \times K$ dimensional regional features 
	into $V$ regions 
	such that $o_u$ in (\ref{eq:6}) represents the $u^{th}$ region of $C^L$ codebook. 
	
	\vspace{-3mm}
	\begin{equation}C^L =\{ o_u \ \forall  \ u \in \{1,...,V\}\}, \  V \in \{64,128,256\} \label{eq:6}\vspace{-1mm}\end{equation}
	
	Using the learned codebook, $F^L$ regions of benchmark test / reference traverses are quantized in (\ref{eq:7}) to predict the clusters or labels $Z^L$, 
	where $\alpha$ is the quantization function. Employing regions-based $F^L$ feature, predicted labels $Z^L$ and regional codebook $C^L$, summed residue $v$ corresponding to each $u^{th}$ region can be retrieved using (\ref{eq:8}).

	\vspace{-3.5mm}
	\begin{equation}Z^L = \alpha(F^L)\label{eq:7}\vspace{-1mm}\end{equation}
	
	In (\ref{eq:8}), for all the $F^L$ regional features that fall in $u^{th}$ region of the $C^L$ codebook, the residues of $F^L_u$ regions and $C^L_u$ codebook's region center are summed. Sometimes, few regions/words appear more frequently in an image than the statistical expectation known as visual word burstiness \cite{jegou2009burstiness}. Standard techniques include power normalization \cite{do2018selective} is performed in (\ref{eq:9}) to avoid it where each $1 \times K$ dimensional residue $v_u$ undergoes non-linear transformation $\gamma$. In (\ref{eq:10}), power normalization is followed by $l_2$ normalization. For each image, $l_2$ normalized residues corresponding to $V$ regions are stored in (\ref{eq:11}) to get final $V \times K$ dimensional VLAD representation $S^L$.
	\vspace{-1mm}
	\begin{equation}v_u = \sum\limits_{F^L_u:Z^L_u=C_u^L} F^L_u-C^L_u\label{eq:8}\vspace{-1mm}\end{equation}
	
	\vspace{-1mm}
	\begin{equation}v_u :=sign(v_u){\|v_u\|}^\gamma\label{eq:9}\vspace{-1mm}\end{equation}
	\begin{equation}\vspace{-2mm}v_u :=\frac{v_u}{\sqrt{v^T_uv_u}}\label{eq:10}\end{equation}
	\vspace{-1mm}
	\begin{equation}S^L = \{v_u \ \forall  \ u \in \{1,...,V\}\}\label{eq:11}\vspace{-1mm}\end{equation}
	
	To match a test image ``A" against the reference image ``B" in (\ref{eq:12}), the dot/scalar product of their $u^{th}$ regional VLAD components $S^{L^A}_u$ and $S^{L^B}_u$, each with dimension $1 \times K$ reaches to an individual regional matching score $j^{A,B}_u$, as visualized in Figure \ref{figure:proposedMethod} (h).
	\vspace{-1mm}
	\begin{equation}j^{A,B}_u =\frac{(S^{L^A}_u) .(S^{L^B}_u)}{\|(S^{L^A}_u)\|\|(S^{L^B}_u)\|}\label{eq:12}\end{equation}
	
	All the scalar $j^{A,B}_u$ scores for $V$ regions are summed up in (\ref{eq:13}) to get final single $J^{A,B}$ matching score. For each test image ``A", the cosine matching in (\ref{eq:12}) is performed against all the reference images and finally, reference image ``X" with the highest similarity score is picked as a matched image using (\ref{eq:14}).
	\vspace{-1.5mm}
	\begin{equation}J^{A,B} =\sum\limits_{u=1}^Vj^{A,B}_u\vspace{-1mm}\label{eq:13}\end{equation}
	\begin{equation}\vspace{-0.5mm}P^A =\argmax_X J^{A,X}\vspace{-1mm}\label{eq:14}\end{equation}

	\vspace{-2mm}
	\section{Datasets, Implementation details, Results and Analysis}
	This section presents the implementation details of our proposed system which will attempt to evaluate its run-time performance for real-time robotic VPR applications. Comparison of the proposed method with state-of-the-art VPR and image retrieval algorithms has been conducted over several benchmark datasets and the obtained results are stated. The section ends by displaying the results on correctly matched and mismatched scenarios of our proposed Region-VLAD framework along with a discussion on the same.
	
	\vspace{-2mm}
	\subsection{Benchmark Place Recognition Datasets}
	More specifically, challenging benchmark VPR datasets \textit{Berlin A100}, \textit{Berlin Halenseestrasse} and \textit{Berlin Kudamm} (see \cite{chen2017only} for detailed introduction), collected from crowd-sourced geotagged photo-mapping platform \textit{Mapillary} are used to evaluate the proposed VPR framework. Each dataset covers two traverses of the same route uploaded by different users. One traverse is used as $R$ reference traverse and the other traverse is employed as $T$ test traverse (see TABLE \ref{table:datasets}). $R'$ represents the reduced reference traverse which matches with $T'$ test traverse (discussed in section \ref{section:MatchScoreThreshold}). Another dataset, \textit{Gardens Point} was captured at QUT campus with one traverse taken in daytime on left side walk and the other traverse was recorded in right side walk at night time \cite{chen2017deep}. The \textit{Synthesized Nordland} dataset was recorded on a train journey with one traverse taken in winter and the other traverse was recorded in spring. Viewpoint variance was added by cropping frames of summer traverse to keep 75\% resemblance \cite{chen2018learning}. 
	For \textit{Berlin A100}, \textit{Berlin Halenseestrasse} and \textit{Berlin Kudamm}, geotagged information is used for ground truth with $0$ to ${\pm2}$ frame tolerance. 
	For \textit{Gardens Point} and \textit{Synthesized Nordland}, the ground truth data is obtained by parsing the frames and maintaining place level resemblance with $0$ to ${\pm3}$ and $0$ to ${\pm2}$ frame tolerance.

\renewcommand{\arraystretch}{0.7}
\renewcommand{\tabcolsep}{0.85pt}
\begin{table}[ht]
	\vspace{-8mm}
	\caption{Benchmark place recognition datasets}
	\label{table:datasets}
\begin{tabular}{|c|c|c|c|c|c|c|c|}
\hline
\multirow{2}{*}{\textbf{Dataset}} & \multirow{2}{*}{\textbf{Environment}} & \multicolumn{2}{c|}{\textbf{Variation}} & \multirow{2}{*}{\textbf{T}} & \multirow{2}{*}{\textbf{R}} & \multirow{2}{*}{\textbf{T'}} & \multirow{2}{*}{\textbf{R'}} \\ \cline{3-4}
 &  & \textbf{Viewpoint} & \textbf{Condition} &  &  &  &  \\ \hline
Berlin A100 & urban & moderate & moderate & 81 & 85 & 70 & 64 \\ \hline
\begin{tabular}[c]{@{}c@{}}Berlin \\ Haleenseetrasse\end{tabular} & \begin{tabular}[c]{@{}c@{}}urban, \\ suburban\end{tabular} & very strong & moderate & 67 & 157 & 50 & 138 \\ \hline
Berlin Kudamm & urban & very strong & moderate & 222 & 201 & 166 & 151 \\ \hline
Gardens Point & campus & strong & strong & 200 & 200 & 152 & 150 \\ \hline
\begin{tabular}[c]{@{}c@{}}Synthesized \\ Nordland\end{tabular} & train & moderate & very strong & 1622 & 1622 & 1221 & 1217 \\ \hline
\end{tabular}
\vspace{-6mm}
\end{table}
	\subsection{Setup, Implementation details and Scalability}
	The proposed VPR framework is implemented in Python $3.6.4$ and the system average runtime over 5 iterations is recorded with $1125$ images. 
	AlexNet pre-trained on Places365 dataset is employed as a CNN model for region-based features extraction with $256 \times 256$ input image size. 
	For all the baseline experiments, we utilize middle \textit{conv3} convolutional layer only due to its better performance in various VPR approaches \cite{sunderhauf2015place}\cite{panphattarasap2016visual}.

	For a single image, a forward pass takes around an average $0.305 ms$ using Caffe on NVIDIA P100 and $15.57 ms$ employing Intel Xeon Gold 6134 @3.2GHz. We extract $N$ ROIs with total time comparable with state of the art methods \cite{chen2017only} (see Table \ref{table:comparison}). The VLAD representations are retrieved and matched using $N$ ROIs mapped on $V$ clustered dictionary $C^L$ (trained using $N$ ROIs per image of $2.6k$ dataset). For direct comparison with \cite{chen2017only}, we use $N=200$ with $V=128$. The results are also reported for $N=400$ with $V=256$. Table \ref{table:comparison} shows that for $N=\{200,400\}$ regional settings, our average VLAD matching times are $100x$ and $58x$ faster than \cite{chen2017only}. 
	
\begin{figure}[t]
		\centering
		\vspace{-4mm}
		\includegraphics[width=1.0\linewidth, height=1\linewidth,keepaspectratio]{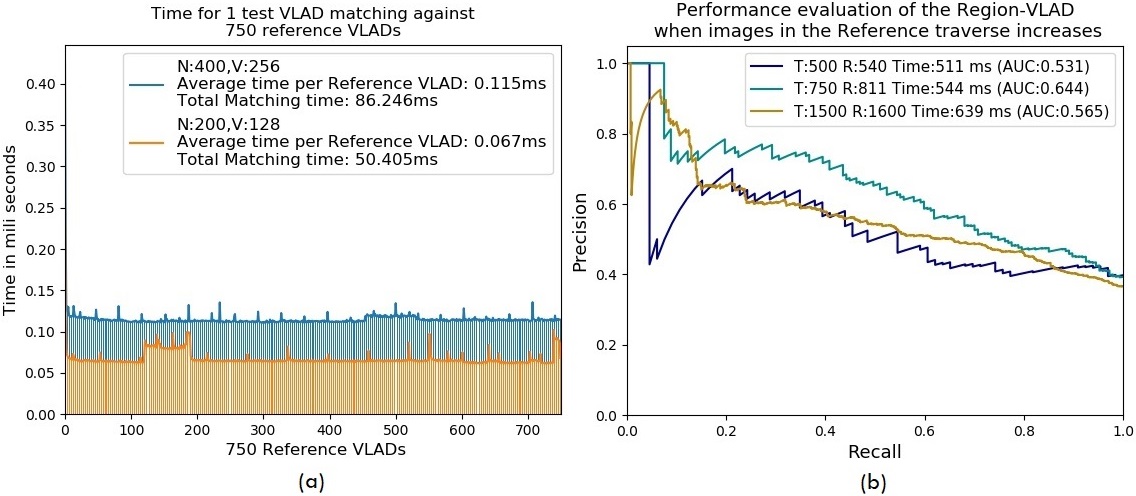}
		\caption{Left (a): Matching times for 1 test VLAD against 750 reference VLADs are presented. Right (b): AUC-PR performance and retrieval time of Region-VLAD are reported while adding more images in T test and R reference traverses.}
		\label{figure:matchTime}
		\vspace{-4mm}
	\end{figure}

	
	In real-time robotic vision applications which include robotic agricultural devices, autonomous infrastructure, environmental monitoring equipment or other agriculture based use-cases, with exploration of new places, the size of the database can grow unbounded. Therefore, scalability is one of an important factor to be considered \cite{cadena2016past}. Under both the regional settings, employing GPU for forward pass and CPUs for both feature extraction and VLAD encoding, the overall times for retrieving a single query VLAD are $396 ms$ and $447 ms$. Whereas, Titan X Pascal GPU in \cite{chen2017only} takes $408 ms$ for feature encoding per query. Figure \ref{figure:matchTime} (a) further confirms that the proposed system consumes an average $0.07 ms$ ($N=200$) and $0.12 ms$ ($N=400$) for matching VLAD representations of a single query and reference image. Therefore, the total retrieval times per query against $R=750$ reference images are approximately around $446.405 ms$ and $533.245 ms$. In comparison, Cross-Region-BoW \cite{chen2017only} takes $7 ms$ for matching features of one test and one reference image. The overall retrieval time against $R=750$ reference images is $5.658 s$ which is $12x$ and $11x$ more than our proposed approaches and practically inappropriate for real-time applications. Our Region-VLAD VPR technique can store the encoded VLAD representations of all the reference frames whereas Cross-Region-BoW needs to perform run-time cross matching of given query regions against all the reference frames' regions, and mutually matched regional features are picked.

	Furthermore, Figure \ref{figure:matchTime} (b) evaluates our proposed system's run-time performance when more places are added in test and reference traverses. For each PR-curve, we employed $T$ test and $R$ reference images. Their VLAD representations are retrieved followed up by their cosine matching and in parallel, we record down the system's performance. We can see that as the size of test and reference traverses increases, the AUC under PR curves remains higher where ``Time" represents the overall matching period for a single test image against $R$ reference traverse. This mimics that the system is capable enough to handle large number of reference/database images while maintaining performance both in terms of accuracy and retrieval time. It should be noted that \cite{chen2017only} used MATLAB implementation which is practically slower than Python but we have employed CPUs in comparison to \cite{chen2017only} which used GPU. 

	\vspace{-6mm}
	\subsection{Comparison Methods}
	To show the dominance of our novel place-centric regions finding approach coupled with VLAD encoding, we replaced VGG-16 with AlexNet365 in \cite{chen2017only} (open-source MATLAB code can be found at \cite{chen2017onlyCode}), and combined the regional features with VLAD and BoW encodings, named as Cross-Region-VLAD and Cross-Region-BoW. For a fair comparison, using $2.6k$ dataset, we trained a separate regional vocabulary employing \textit{conv4} for regions identification and \textit{conv3} for feature extraction. Keeping $N=200$, we used $V=128$ for Cross-Region-VLAD and $V=2.6k$ for Cross-Region-BoW. Furthermore, results are also reported for HybridNet with Spatial Pyramidal Pooling (SPP) \cite{chen2017deep} employed on \textit{conv5} of the model. We also integrated RMAC \cite{tolias2015particular} on AlexNet365  while performing power- and l2-normalization on the retrieved regional features. Similar to \cite{chen2017only}, mutual regions are filtered using cross matching, their scores are summed up and maximum matching score is considered for retrieval.

	PR-curves across all other image retrieval approaches including Cross-Pool, Max-Pool, Sum-Pool, Whole and state-of-the-art VPR approaches FABMAP and SEQSLAM are taken from \cite{chen2017only}. Authors in \cite{chen2017only} employed \textit{conv5\_2} of deep object centric VGG-16 as features representation. However, Cross-Region-BoW \cite{chen2017only} with deep VGG-16 model used \textit{conv5\_3} for landmarks identification and \textit{conv5\_2} for feature extraction. Standard FABMAP implementation \cite{cummins2008fabCode} and three sequential frames configuration for SEQSLAM were used by \cite{chen2017only}.

	
	\renewcommand{\arraystretch}{1}
	\renewcommand{\tabcolsep}{0.9pt}
	
	\begin{table*}[]
		\vspace{-6mm}
		\caption{Comparison of our proposed method with Cross-Region-BoW\cite{chen2017only} }
		\label{table:comparison}
		\begin{tabular}{|c|c|c|c|c|c|c|c|c|c|c|c|c|c|c|c|c|c|c|c|c|}
			\hline
			\multicolumn{2}{|c|}{\textbf{Methodology}}                                                                         & \multicolumn{18}{c|}{Our Region-VLAD}                                                                                                                                       & Cross-Region-BoW \cite{chen2017only}        \\ \hline
			\multicolumn{2}{|c|}{\textbf{Model}}                                                                               & \multicolumn{18}{c|}{AlexNet365}                                                                                                                                            & VGG16             \\ \hline
			\multicolumn{2}{|c|}{\textbf{Images}}                                                                              & \multicolumn{18}{c|}{1125}                                                                                                                                                  & 1000               \\ \hline

			
			\multicolumn{2}{|c|}{\textbf{GPU/CPU}}                                                                             & NVIDIA P100  & \multicolumn{17}{c|}{Intel(R) Xeon(R) Gold 6134 CPU @ 3.20GHz with 32 cores, 64GB RAM}                                                                                                           & Titan X Pascal GPU \\ \hline
			
			
			\multicolumn{2}{|c|}{\textbf{Forward pass time (ms)}}                                                    & 0.305     & \multicolumn{17}{c|}{15.574}                                                                                                                              & 59                 \\ \hline

			\multicolumn{2}{|c|}{\textbf{ROIs ``N"}}                                                                    & \multicolumn{3}{c|}{50}    & \multicolumn{3}{c|}{100}   & \multicolumn{3}{c|}{200}   & \multicolumn{3}{c|}{300}   & \multicolumn{3}{c|}{400}   & \multicolumn{3}{c|}{500}   & 200                \\ \hline
			\multicolumn{2}{|c|}{\textbf{Extraction time (s)}}                                                       & \multicolumn{3}{c|}{0.328} & \multicolumn{3}{c|}{0.361} & \multicolumn{3}{c|}{0.394} & \multicolumn{3}{c|}{0.402} & \multicolumn{3}{c|}{0.443} & \multicolumn{3}{c|}{0.452} & 0.349              \\ \hline
			\multicolumn{2}{|c|}{\textbf{Regions ``V"}}                                                                            & 64           & 128  & 256  & 64      & 128     & 256    & 64      & 128     & 256    & 64      & 128     & 256    & 64      & 128     & 256    & 64      & 128     & 256    & 10k Visual words   \\ \hline
			\multirow{2}{*}{\textbf{\begin{tabular}[c]{@{}c@{}}Matching time\\ (ms)\end{tabular}}} & \textbf{VLAD encoding} & 1.33         & 2.05 & 3.58 & 1.55    & 2.28    & 3.79   & 1.91    & 2.4    & 4.03   & 1.99    & 2.68    & 4.28   & 2.13    & 2.96    & 4.54   & 2.36    & 3.16    & 4.75   & \multirow{2}{*}{7} \\ \cline{2-20}
			& \textbf{VLAD matching} & 0.05         & 0.06 & 0.12 & 0.05    & 0.08    & 0.13   & 0.05    & 0.07    & 0.12   & 0.04    & 0.07    & 0.12   & 0.05    & 0.08    & 0.12   & 0.05    & 0.07    & 0.12   &                    \\ \hline
		\end{tabular}
		\vspace{-2mm}
	\end{table*}
	
	\vspace{-4mm}
	\subsection{Precision Recall Characteristics}
	\label{section:pr_curves}
	In image retrieval tasks where there is a moderate to large class imbalance which means the positive class samples are quite rare as compared to the negative classes, Precision-Recall curves are usually employed as evaluation metric \cite{hanley1982meaning}. 
    For all the benchmark datasets, we first calculate the difference in AUC-PR performance of \cite{chen2017only} and Region-VLAD, determine their average which comes around an overall $13\%$ performance improvement.
    
	\subsubsection{\textit{Berlin Halenseestrasse}}
	
\begin{figure}[h]
		\centering
		\vspace{-4mm}
		\includegraphics[width=1\linewidth, height=0.53\linewidth,keepaspectratio]{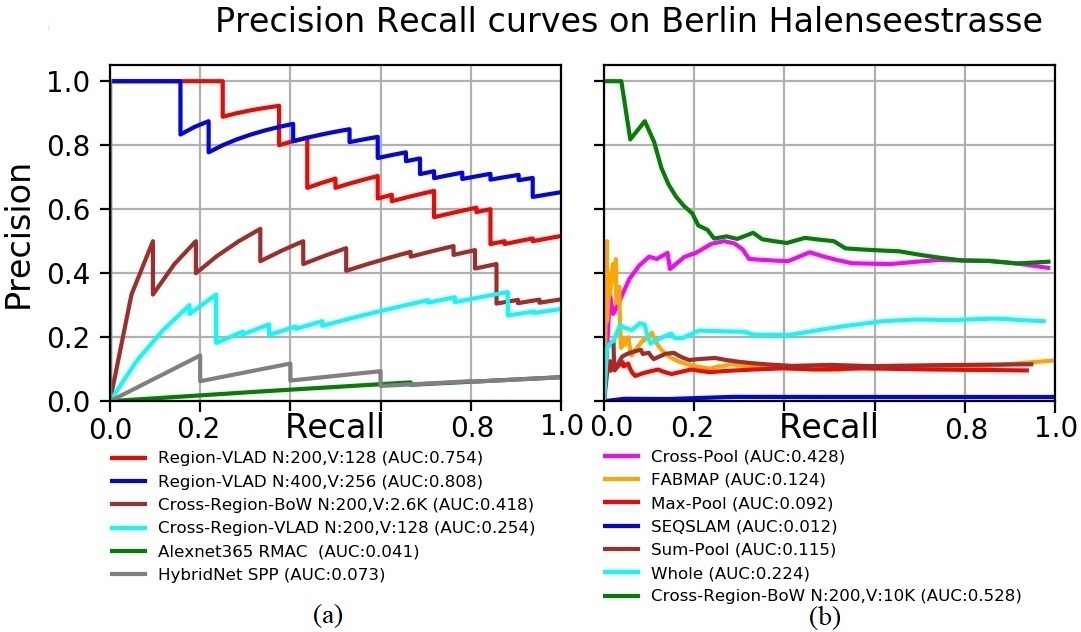}
		\caption{ AUC PR-curves for Berlin Halenseestrasse dataset are presented here. Left: PR-curves of our proposed Region-VLAD and \cite{chen2017only} employed on AlexNet365 with VLAD and BoW encodings. Right: Comparison with state-of-the-art VPR approaches employing VGG-16.}
		\label{figure:berlin_hass}
		\vspace{-2.5mm}
	\end{figure}

	In Figure \ref{figure:berlin_hass} (a), the proposed Region-VLAD PR-curves for \textit{Berlin Halenseestrasse} dataset significantly outperforms all other state-of-the-art methods. Surprisingly, Cross-Region-VLAD PR-curve underperformed with a big margin. This mimics that the better AUC-PR performance of our proposed approach is encouraged with the use of our novel regional features.  
	Furthermore, investigations on Cross-Region-VLAD suggest that under strong viewpoint change, the mapping of cross-convoluted regional patterns  \cite{chen2017only} 
	over the vocabulary for VLAD retrieval results into non-uniform feature distribution. Although, normalization is carried out but still, many zero regional residues exist in the VLAD representation which reflects on the PR-curves. Cross-Region-BoW only considers the mutually matched regions and exhibits better results. Moreover, RMAC which is state-of-the-art in other image retrieval techniques and SPP, both are sensitive under strong viewpoint variation, thus under-performed on this dataset.
	
	Although, FABMAP is robust under viewpoint variation but it still underperformed on this dataset just like SEQSLAM, a whole image-based technique which subtracts patch-normalized sequence of frames. Cross-Pool employs a similar idea of pooling as Cross-Region-BoW, so both have achieved nearly the similar PR-curves whereas other pooling techniques under-performed.  
	It is worth noting that even with smaller regional dictionaries, our proposed Region-VLAD framework still achieves better results than VGG-16 based Cross-Region-BoW \cite{chen2017only} and other methodologies. It highlights the potential of our shallow CNN based regional features robustness under strong viewpoint variations.

	\subsubsection{\textit{Berlin Kudamm}}
\begin{figure}[h]
		\centering
		\vspace{-4mm}
		\includegraphics[width=1\linewidth, height=0.53\linewidth,keepaspectratio]{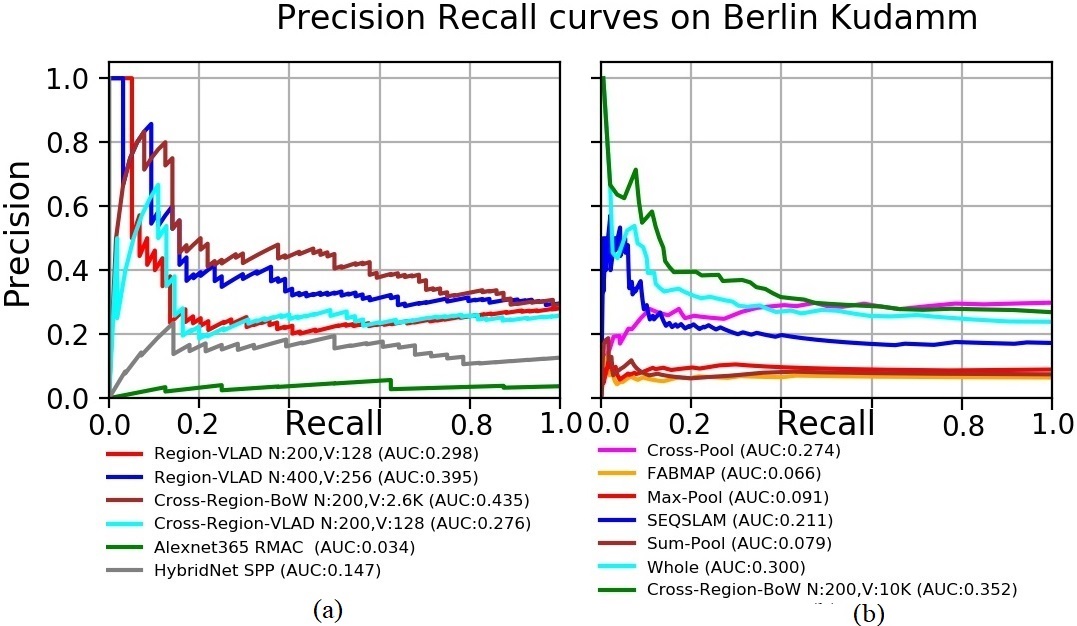}
		\caption{AUC PR-curves for Berlin Kudamm dataset are presented here. Left: PR-curves of our proposed Region-VLAD and \cite{chen2017only} employed on AlexNet365 with VLAD and BoW encodings. Right: Comparison with state-of-the-art VPR approaches employing VGG-16.}
		\label{figure:berlin_kud}
		\vspace{-3mm}
	\end{figure}

	Due to urban environment, too many dynamic and confusing objects such as vehicles, trees and pedestrians with homogeneous scenes lead to perceptual aliasing coupled with severe viewpoint changes makes it a challenging dataset. 
    Figure \ref{figure:berlin_kud} (a) shows that our proposed Region-VLAD approach still manages to achieve better results. AlexNet365 combined with Cross-Region-BoW claims state-of-the-art results with $V=2.6k$ regional vocabulary. 
    RMAC and SPP again underperformed. This is apparently because VPR is different from other image retrieval and recognition systems where a single object majorly covers the whole image. Therefore, Sum-Pool, Max-Pool and RMAC which perform relatively well in such vision-based tasks actually not performed well in VPR under strong viewpoint and appearance changes. 
    
    In Figure \ref{figure:berlin_kud} (b), due to resemblance among the places captured in sequence, Whole and SeqSLAM with their whole-image based approach have shown better performances. 
    With higher precision at start and as recall increases, Region-VLAD PR-curves are quite similar but covering more AUC than Whole, SeqSLAM, Cross-Pool and VGG-16 Cross-Region-BoW. 

	\subsubsection{\textit{Berlin A100}}
\begin{figure}[h]
		\centering
		\vspace{-3.5mm}
		\includegraphics[width=1\linewidth, height=0.53\linewidth,keepaspectratio]{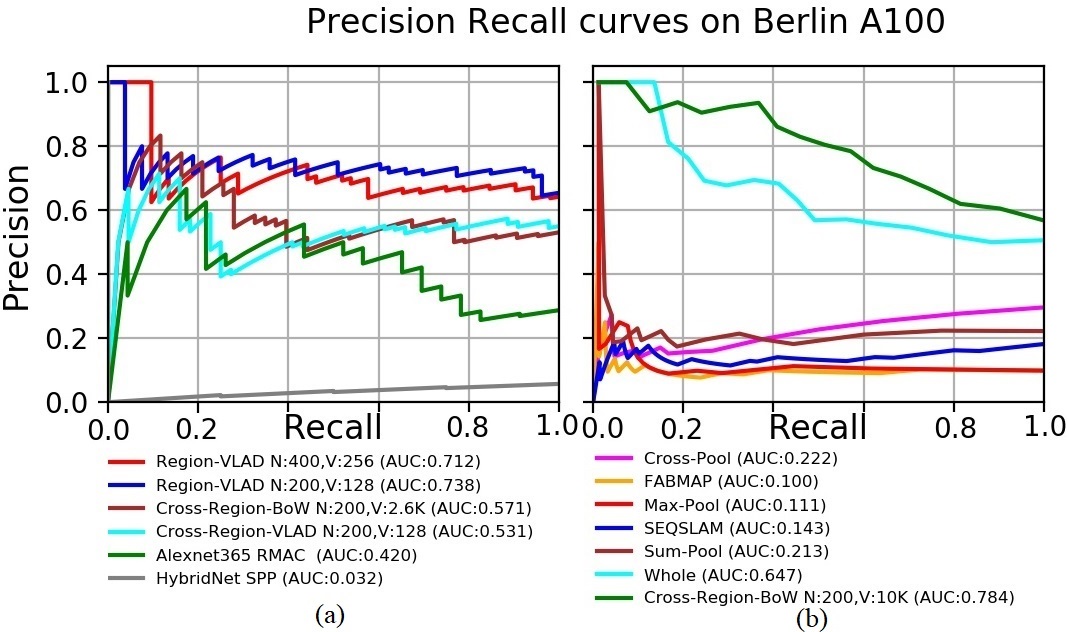}
		\caption{AUC PR-curves for Berlin A100 dataset are presented here. Left: PR-curves of our proposed Region-VLAD and \cite{chen2017only} employed on AlexNet365 with VLAD and BoW encodings. Right: Comparison with state-of-the-art VPR approaches employing VGG-16.}
		\label{figure:berlin_a100}
		\vspace{-3mm}
		
	\end{figure}

	This dataset exhibits moderate viewpoint and moderate conditional changes coupled with dynamic objects. PR-curves are displayed in Figure \ref{figure:berlin_a100}. 
	It is quite evident that our Region-VLAD approach in Figure \ref{figure:berlin_a100} (a) achieves similar results as state-of-the-art VGG-16 Cross-Region-BoW \cite{chen2017only}. AlexNet365 combined with cross-regional approach of \cite{chen2017only} achieves similar and better results for BoW and VLAD. SPP employed on HyridNet was found not very convincing. It might be because HybridNet is fine-tuned on SPED which contains minimal dynamic instances among the same place(s) captured over multiple times of the year. 
	
	Against our approach, RMAC on AlexNet365 achieves comparable and better performance than FABMAP and pooling techniques including Sum-Pool, Max-Pool and Cross-Pool. Since condition and viewpoint variations are not much stronger in this dataset, therefore, RMAC and other approaches have also shown better performance. A deep analysis on the dataset reveals varied time interval between the captured frames due to which SEQSLAM underperformed on this dataset. Overall, our proposed Region-VLAD achieved second best performance after VGG-16 Cross-Region-BoW \cite{chen2017only}.    
	
	\subsubsection{\textit{Synthesized Nordland}}
\begin{figure}[t]
		\centering
		\vspace{-3.5mm}
		\includegraphics[width=1\linewidth, height=0.53\linewidth,keepaspectratio]{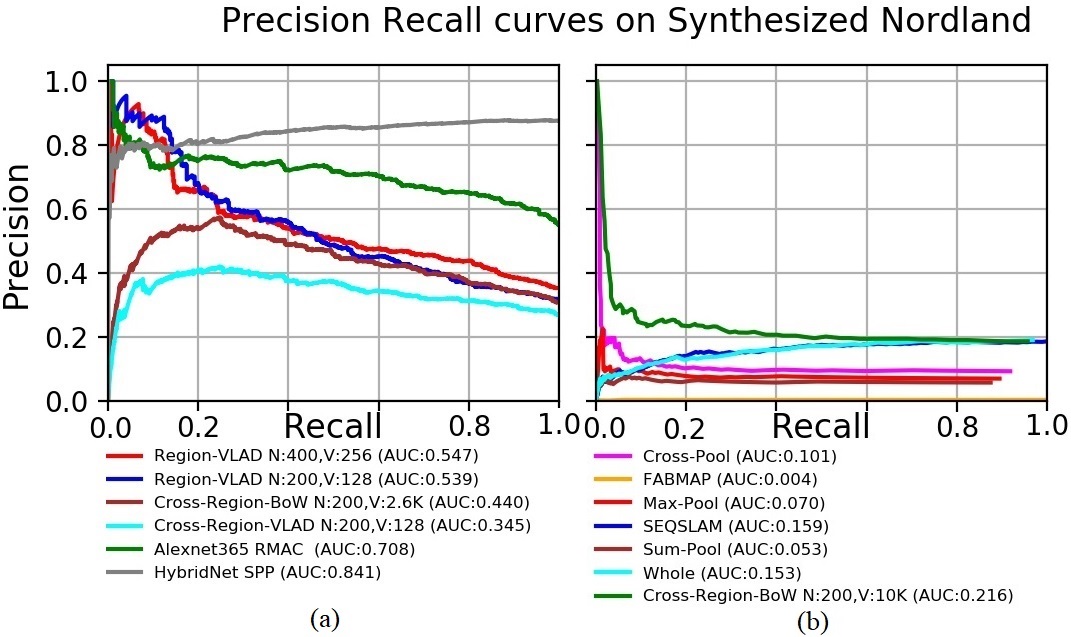}
		\caption{AUC PR-curves for Synthesized Nordland dataset are presented here. Left: PR-curves of our proposed Region-VLAD and \cite{chen2017only} employed on AlexNet365 with VLAD and BoW encodings. Right: Comparison with state-of-the-art VPR approaches employing VGG-16.}
		\label{figure:syn_nordland}
		\vspace{-4.5mm}
	\end{figure}

	In comparison to other approaches, PR-curves in Figure \ref{figure:syn_nordland} (a) show that our proposed approach works relatively well on this dataset but RMAC and SPP achieve state-of-the-art performances. Employing deep VGG-16, Max-Pool and Sum-Pool have not shown better results and similar 
	whole image-based techniques i.e. SEQSLAM and Whole exhibit similar PR-curves. 
	
	
	
	Since in HybridNet, fine-tuning the CNN model with SPED induced condition invariance. 
	Thus, employing SPP on HybridNet has shown superior performance on this dataset (exhibiting strong conditional changes). In comparison, scene-centric AlexNet365 integrated with Cross-Region-BoW and Cross-Region-VLAD outperformed deep ImageNet-centric VGG-16 based Cross-Region-BoW \cite{chen2017only}. This highlights the importance of CNN training. 

    \subsubsection{\textit{Gardens Point}}	
\begin{figure}[h]
		\centering
		\vspace{-4.2mm}
		\includegraphics[width=1\linewidth, height=0.53\linewidth,keepaspectratio]{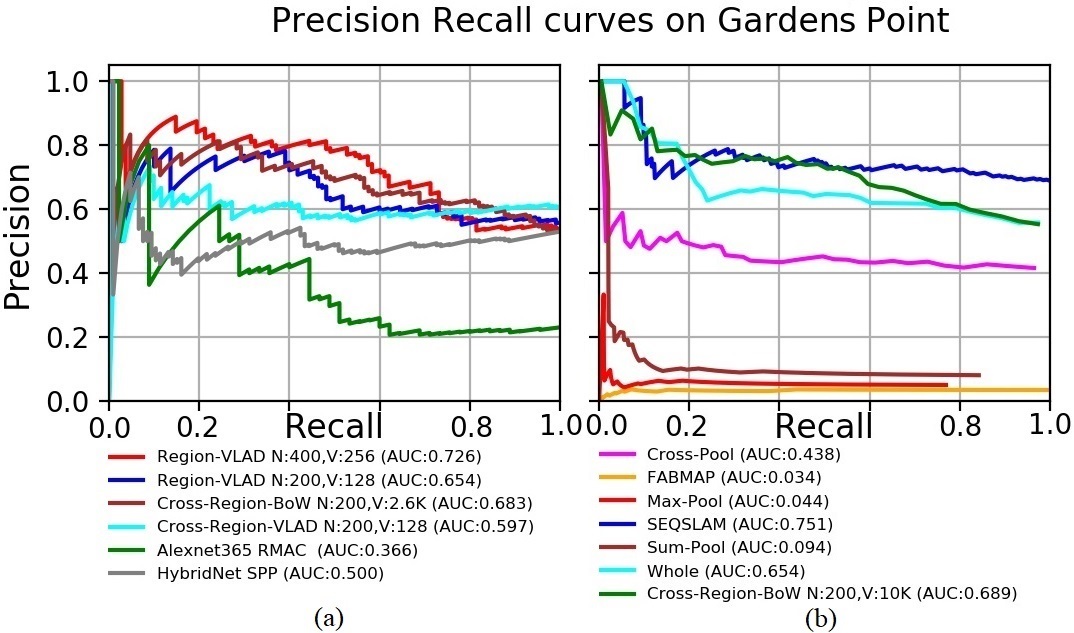}
		\caption{AUC PR-curves for Gardens Point dataset are shown here. Left: PR-curves of our proposed Region-VLAD and \cite{chen2017only} employed on AlexNet365 with VLAD and BoW encodings. Right: Comparison with state-of-the-art VPR approaches.}
		\label{figure:garden_point}
		\vspace{-3mm}
	\end{figure}

	Both the \textit{Gardens Point} traverses exhibit stronger lightning variations with adequate temporal coherence between the frames. Figure \ref{figure:garden_point} shows that our Region-VLAD approach achieves similar and better performance than Cross-Region-BoW, Cross-Region-VLAD, Whole, RMAC and SPP. Taking advantage from the sequential information, SEQSLAM has shown state-of-the-art performance. Cross-Region-BoW and Cross-Region-VLAD integrated with AlexNet and VGG-16 exhibit similar performances but approaches including Sum-Pool, Max-Pool and FABMAP relatively underperformed. 
	

	
	\vspace{-4.5mm}


	\subsection{Matching Score Thresholding}\label{section:MatchScoreThreshold}
	
\begin{figure}[t]
		\centering
		\vspace{-4mm}
		\includegraphics[width=0.95\linewidth, height=0.66\linewidth]{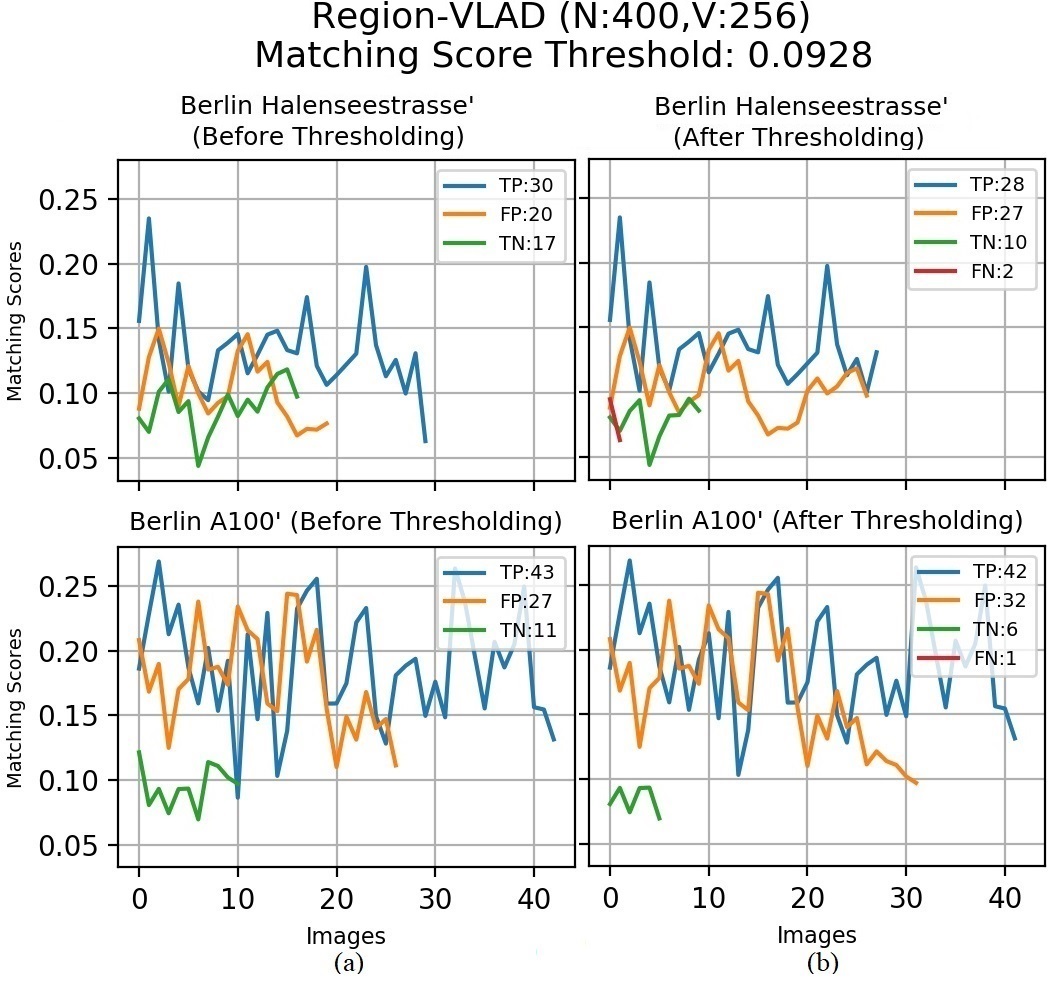}
		\caption{Left column presents graphs for \textit{Berlin Halenseestrasse} and \textit{Berlin A100} before thresholding and right column graphs showcase the change in TP, FP, TN and FN upon thresholding. Our proposed Region-VLAD framework assigned low score to the T-T' or TN queries.}
		\label{figure:tp_fp_fn_vlad}
		\vspace{-4.5mm}
	\end{figure}

	By nature, PR curves do not consider True Negative cases (correctly missed the non existing events/classes) \cite{hanley1982meaning}. So, in order to tackle such tricky situations, we employ $T$ test traverse and $R'$ reference traverse from all the datasets so that $T-T'$ queries can be treated as new places (see Table \ref{table:datasets}).	
	Figure \ref{figure:tp_fp_fn_vlad} visualizes the results of the proposed Region-VLAD framework before and after the match score thresholding. On the basis of matching scores, y-axis differentiates the TP, FN, FP and TN events\footnote{TP for True Positive, FN for False Negative, FP for False Positive and TN for True Negative} shown with different coloured curves, where length of the curves in x-axis denotes the number of images which the events contain. The threshold is an average of TN scores of $R'$ reference traverses of the benchmark datasets. Due to limited space, results are reported for two datasets only. Upon thresholding in Figure \ref{figure:tp_fp_fn_vlad}(b), Region-VLAD for \textit{Berlin Halenseestrasse} dataset missed $FN=2$ correctly matched images and successfully filtered $10$ queries out of $TN=17$. The same behavior is observed for \textit{Berlin A100} dataset. 
	In scenarios when the system comes across previously observed places as well as new places, it becomes increasingly challenging to successfully retrieve the correct matches (TPs), discard incorrect matches (TNs), while reducing FPs (retrieved incorrect matches) and FNs (discarded correctly retrieved matches). It is evident that Region-VLAD not only boosts up the AUC under PR-curves but also deals efficiently in assigning low scores to TN queries (green curves). 

	\vspace{-3mm}
	
	\subsection{Analysis}

\begin{figure}[h]
		\centering
		\vspace{-2mm}
		\includegraphics[width=1\linewidth, height=1\linewidth,keepaspectratio]{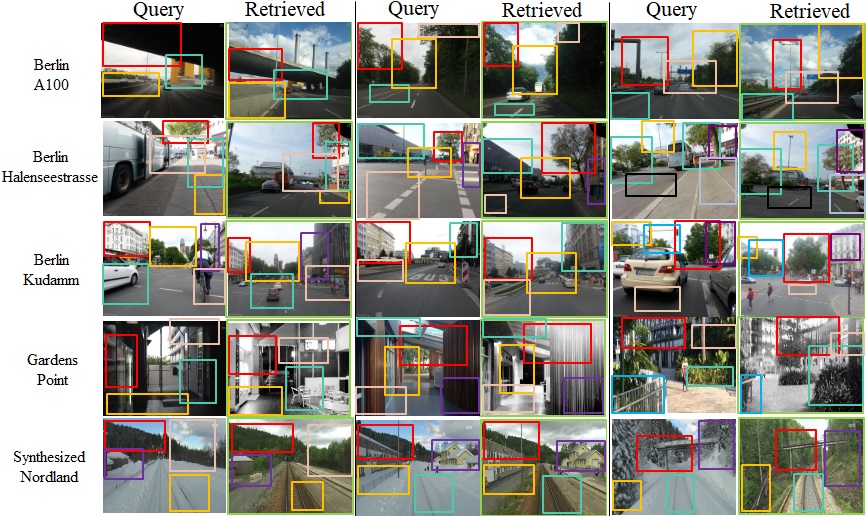}
		\caption{Sample correctly retrieved matches using the proposed VPR framework are presented here; it identifies common regions across the queries and retrieved images under strong viewpoint and appearance variations.}
		\label{figure:match}
		\vspace{-2mm}
	\end{figure}

\begin{figure}[h]
		\centering
		\vspace{-3mm}
		\includegraphics[width=1\linewidth, height=0.55\linewidth,keepaspectratio]{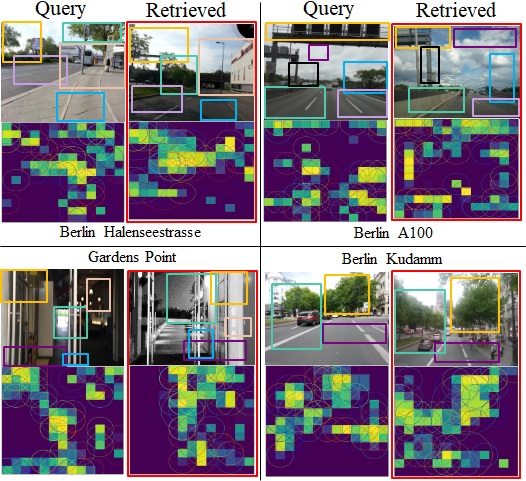}
		\caption{Sample incorrectly retrieved matches using the proposed VPR framework are presented here; each query and the retrieved database images are geographically different but exhibiting similar scenes and conditions.}
		\label{figure:unmatch}
	\end{figure}

	Figure \ref{figure:match} and Figure \ref{figure:unmatch} illustrate some of the matched and mismatched scenarios. For the correct matches, taking advantage from CNN's scene-centric training, Region-VLAD identifies the common regions shown with different coloured boxes under simultaneous viewpoint and appearance changes. For the mismatched scenarios, the identified top novel regions with coloured boxes (trees, lamp posts) show the areas where the system is interested in and matches the scenes but wrongly recognizes the places. 
    We have seen that Cross-Region-BoW \cite{chen2017only} when integrated with AlexNet365 showed comparable performance but at high time computation cost. However, our Region-VLAD still outperformed Cross-Region-BoW \cite{chen2017only} with smaller dictionary and low retrieval time. Also, cross-regional approach of \cite{chen2017only} when combined with the VLAD shown inferior results which confirms the performance boost in Region-VLAD encouraged with our novel regional approach. 
    Datasets and results are placed at \cite{ahmedest61VPR} and the author intends to open-source the code upon publication.

	\vspace{-3mm}
	
	\section{Conclusion}
	
	For Visual Place Recognition on resource-constrained mobile robots, achieving state-of-the-art performance/accuracy with lightweight CNN architectures is highly desirable but a challenging problem. This paper has taken a step in this direction and presented a holistic approach targeted for a CNN architecture comprising a small number of layers pre-trained on a scene-centric image database to reduce the memory and computational cost for resource-constrained mobile robots. The proposed framework detects novel CNN-based regional features and combines them with the VLAD encoding methodology adapted specifically for computation-efficient and environment Invariant-VPR problem. The proposed method achieved state-of-the-art AUC-PR curves on significant viewpoint- and condition-variant benchmark place recognition datasets.
	
	\balance
	In future, it would be useful to analyse the performance of the proposed framework on other shallow/deep CNN models individually trained/fine-tuned on place recognition-centric datasets. Furthermore, instead of employing defined number of novel regions, it would be interesting to investigate the dynamic regional features selection at runtime and their performances on multiple regional vocabularies. 
	
	\vspace{-9mm}
	
	


	
	
	.

	
	\bibliographystyle{IEEEtran}
	\bibliography{root}

\end{document}